Preprint from: **https://www.gipp.com/pub/**

T.Spinde, K. Sinha, N. Meuschke, B. Gipp, "TASSY—A Text Annotation Survey System", in Proceedings of the ACM/IEEE Joint Conference on Digital Libraries (JCDL), 2021.

# TASSY—A Text Annotation Survey System

Timo Spinde
*University of Konstanz*
Konstanz, Germany
Timo.Spinde@uni-konstanz.de

Kanishka Sinha
*University of Passau*
Passau, Germany
sinha02@ads.uni-passau.de

Norman Meuschke
*University of Wuppertal*
Wuppertal, Germany
meuschke@uni-wuppertal.de

Bela Gipp
*University of Wuppertal*
Wuppertal, Germany
gipp@uni-wuppertal.de

***Abstract*— We present a free and open-source tool for creating web-based surveys that include text annotation tasks. Existing tools offer either text annotation or survey functionality but not both. Combining the two input types is particularly relevant for investigating a reader's perception of a text which also depends on the reader's background, such as age, gender, and education. Our tool caters primarily to the needs of researchers in the Library and Information Sciences, the Social Sciences, and the Humanities who apply Content Analysis to investigate, e.g., media bias, political communication, or fake news.**

## I. Introduction

Text annotation, i.e., reading text and assigning labels to parts thereof, is an essential task in several disciplines. Natural Language Processing (NLP) research heavily relies on text annotation to create datasets for training and evaluating automated approaches for, e.g., parts-of-speech tagging, named entity recognition, and dependency parsing [1]–[4]. In Library and Information Science, the Social Sciences, and the Humanities, Content Analysis is an essential research method that involves codified text annotation [5]. The technique has many applications, such as performing authorship analysis or quantifying the public perception of products, events, or brands.

Many text annotation tasks, particularly in NLP, include labeling parts or properties of a text for which an objective truth exists. Examples include tagging parts of speech or resolving coreferences. For such tasks, the characteristics pf the annotators are irrelevant. For Content Analysis tasks that study the perception of text, annotator characteristics, such as gender, education, and political view, are highly relevant. Examples include the analysis of media bias and fake news.

Many tools support text annotation tasks, i.e., aiding human coders in finding relevant parts of a text and assigning labels. Likewise, numerous, typically web-based tools allow creating customized surveys.

We created our tool, TASSY, to address a basic yet significant shortcoming of existing tools—no tools we are aware of offer both text annotation and survey functionality. For a research project on identifying media bias, we needed to task several hundred participants with annotating words and phrases the participants perceived as opinionated in news articles. Moreover, we needed to record personal background information for each participant. We did not find a tool that allowed recording both types of data. Therefore, we developed TASSY as a free and open-source web-based system for creating surveys that include text annotation tasks.

This work was supported by the German Academic Exchange Service (DAAD) and the Hanns-Seidel-Foundation.

## II. Related work

NLP text annotation tools typically offer templates for common tasks like labeling named entities. The templates propose a workflow, GUI layout, and labels annotators should use. The focus is on minimizing the time for creating annotation tasks and maximizing annotator efficiency. To this end, the tools offer search and navigation functions that help users to find relevant text parts. Moreover, the tools employ artificial intelligence to reduce annotation time by recommending labels. Text annotation tools commonly allow customizing the task-specific templates or create new templates from scratch to support other annotation tasks. However, none of the tools offers ready-to-use functionality to create survey questions.

The ability to create, distribute, and analyze large-scale surveys characterizes another class of tools. Survey tools are predominantly web platforms that allow users to create surveys by customizing question templates and configuring the survey logic, e.g., specifying follow-up questions that depend on the answer to a previous question. Despite the typically large number of features in survey tools, we could not find a tool that offered a text annotation question type.

TABLE I summarizes three of the most widely-used text annotation and survey tools to highlight their complementary strengths that triggered the development of our tool, TASSY.

TABLE I. Widely-used text annotation and survey tools.

| | | Text Annotation Tools | | | Survey Tools | | |
|---|---|---|---|---|---|---|---|
| **Criterion** | **TASSY** | **Prodigy** | **Tag-tog** | **Doc-cano** | **SoGo Survey** | **Survey Monkey** | **Lime Survey** |
| Text annotations | ✔ | ✔ | ✔ | ✔ | ✘ | ✘ | ✘ |
| Survey questions | ✔ | ✘ | ✘ | ✘ | ✔ | ✔ | ✔ |
| Delivery | SWA | SWA | Cloud | SWA | Cloud | Cloud | Cloud / SWA |
| Costs | Free | $390–$490 | $0–$99 p.m. | Free | $0–$99 p.m. | €0–€99 p.m. | $0–$849 p.a. |
| License | MIT | Commercial | Commercial | MIT | Commercial | Commercial | GNU |

**Legend:** SWA—Self-hosted web application

## III. System Overview

TASSY is a progressive web application implemented using the Python Flask and Vue.js frameworks that supports a MySQL/MariaDB or SQLite database. The tool's appearance and functionality are highly customizable via a Python API.

The **source code** (MIT license) is available at **https://bit.ly/35GNqOd**. A **demo system** showcasing the survey on media bias that triggered the development of TASSY is available at **http://tassy.blind-review.org**.

Fig. 1 shows a question in TASSY that combines a text annotation task with a single-select input. The order of the questions can be configured to be identical for all participants or chosen randomly. The text that participants shall annotate is given in italic font ①. The question or task description ② can contain instructions ③. If desired, a link to detailed instructions, e.g., a coding book, can be inserted. Clicking the link will open the instructions in a modal window. Participants can select parts of the provided text at will. Ending the selection or pausing it for one second triggers the system to store the selected text as an annotation. Annotated text parts are highlighted in yellow and shown below the instructions ④. Participants can reverse any annotation by clicking the x-button associated with each annotation. Administrators can configure lower and upper bounds on the length of the text parts that users may select. In the demo application, the maximum number of words that participants can select for one annotation is set to six. Exceeding the threshold will show an error overlay, informing the user that this selection is invalid. The lower part of the screen shows a single-select survey input ⑤. Administrators can configure which inputs are mandatory. In the demo application, annotating parts of the text is optional, while completing two single-select questions is mandatory. Note that Fig. 1 only shows one of the two questions. Providing the required inputs activates the next button at the top of the page that allows proceeding to the next question ⑥. A configurable section label and progress indicator inform the participants about the progression of the survey ⑦.

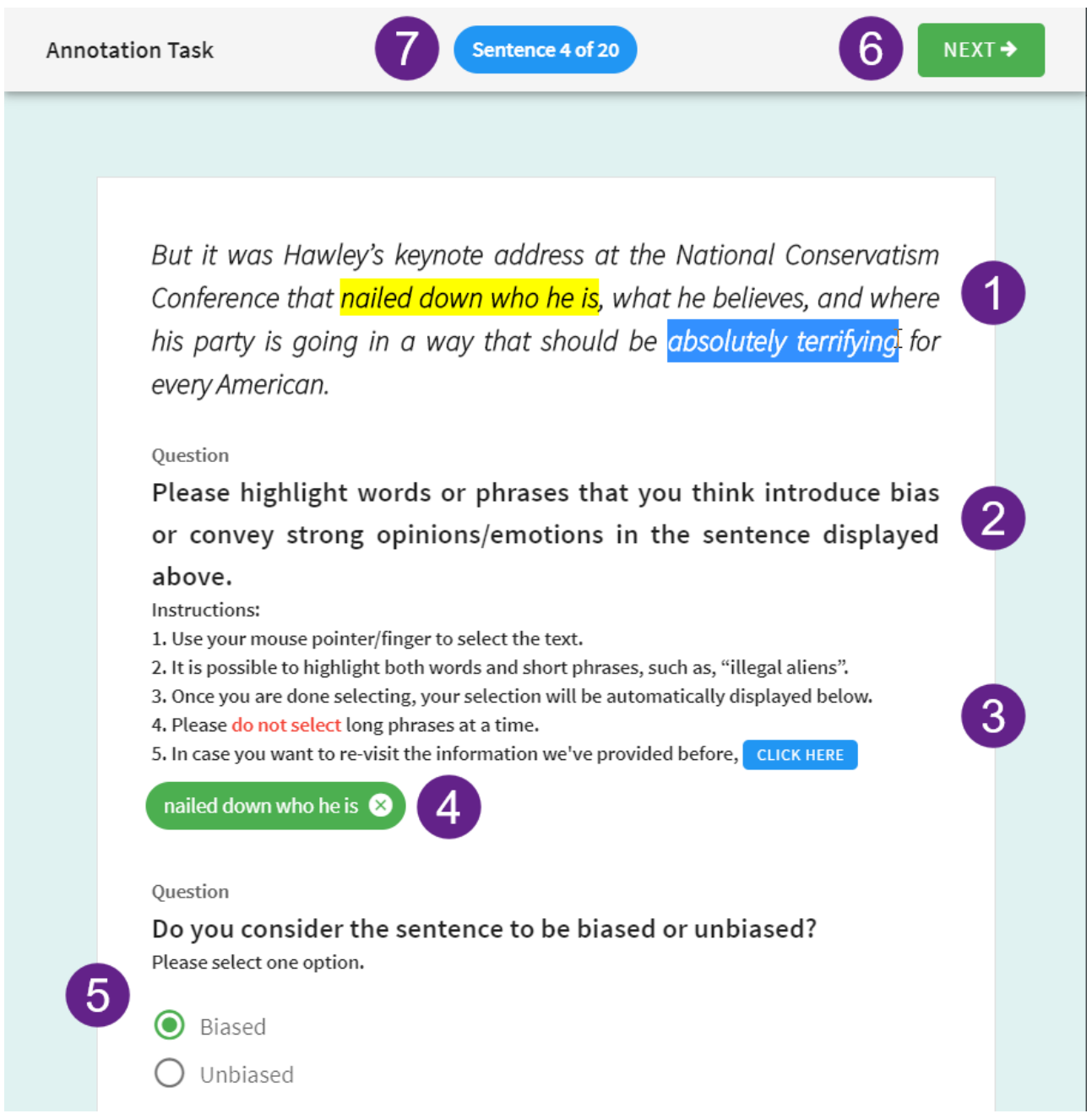

Fig. 1. Survey question requiring text annotation.

The system includes several other input types that allow creating versatile surveys. Fig. 2 exemplifies a simple numeric input to record the age of participants. Fig. 3 shows a slider input that enables participants to indicate their political views on a discretized scale ranging from very liberal to very conservative. TASSY also enables open-ended questions, for which users can input text freely or as an extension to existing answers. Fig. 4 shows an example of the latter. The user can select one or multiple answers from the provided list and optionally enter additional answers for the shown question. For all input types, administrators can configure the inputs required for a) activating the next button that allows proceeding to the next question and b) the submit button that enables completing a section of the survey (for the example shown in Fig. 4, the section on participants' news consumption).

## IV. Future Work

In the future, we plan to enable the customization and administration of text annotation surveys via a graphical user interface rather than a Python API. Furthermore, we plan to offer a free hosted version of TASSY to eliminate the need to install the system locally. The planned extensions will increase the usability for less tech-savvy users.

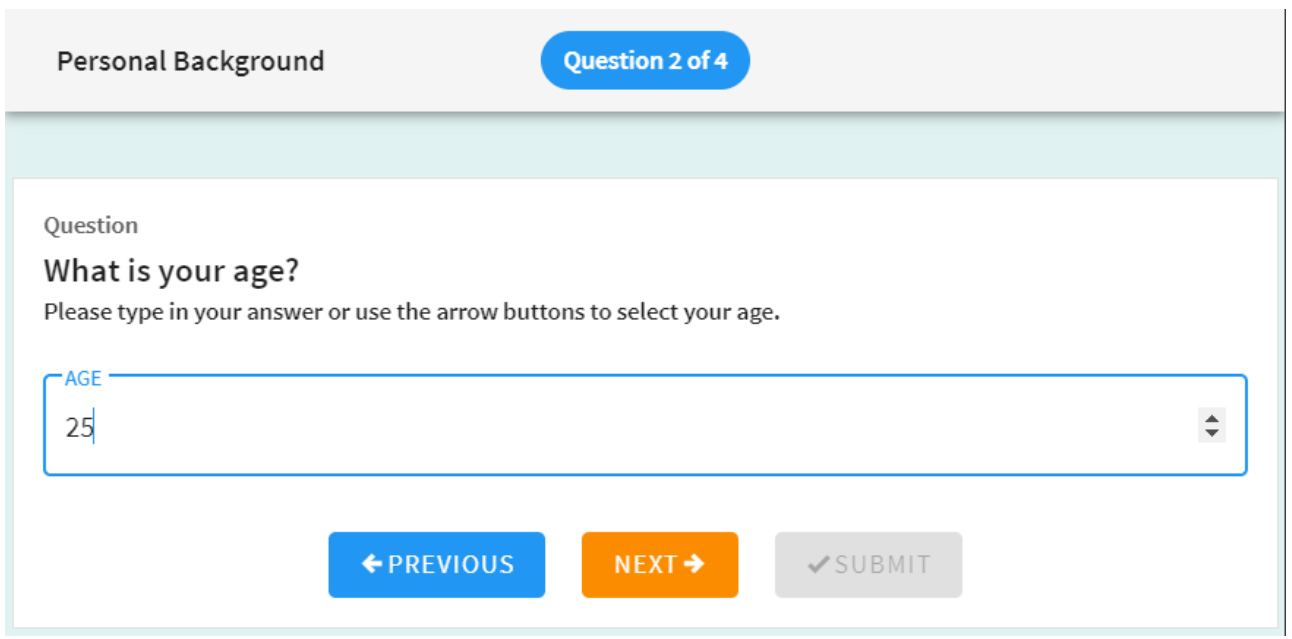

Fig. 2. Example of a numeric input.

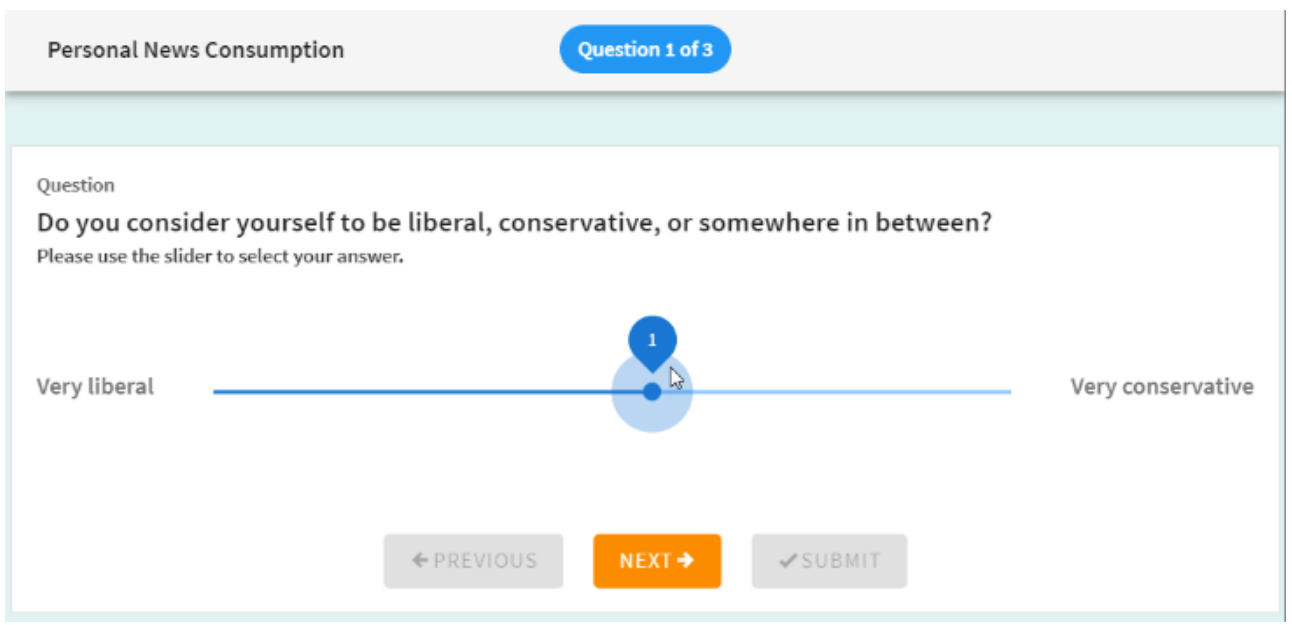

Fig. 3. Example of a slider input.

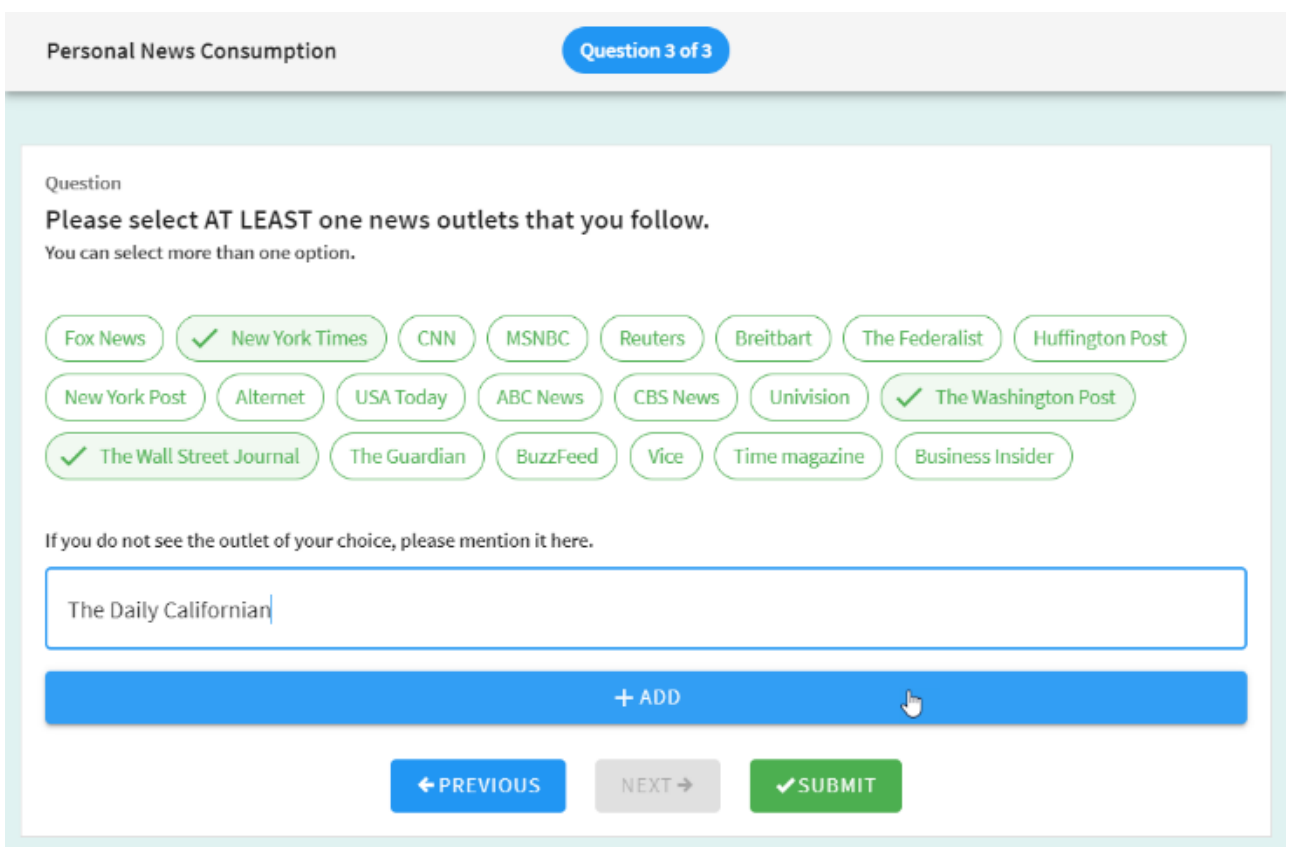

Fig. 4. Example of an extensible multi-select input.

## How to cite this paper:

## BibTex:

```
@InProceedings{Spinde2021c,
  title = {TASSY - A Text Annotation Survey System},
  booktitle = {Proceedings of the {ACM}/{IEEE} {Joint} {Conference} on {Digital} {Libraries} ({JCDL})},
  author = {Spinde, Timo and Sinha, Kanishka and Meuschke, Norman and Gipp, Bela},
  year = {2021},
  month = {Sep.},
  topic = {misc},
}
```